\documentclass{article}
\usepackage{spconf,amsmath,graphicx}
\usepackage{url}
\usepackage{tikz}
\usepackage{amssymb}

\title{Recursive Joint Attention for Audio-Visual Fusion\\ in Regression based Emotion Recognition}
\name{R Gnana Praveen, Eric Granger and Patrick Cardinal \thanks{This research was supported by the Natural Sciences and Engineering Research Council of Canada, and the Fonds de recherche du Québec – Nature et technologies.}}
\address{Laboratoire d’imagerie, de vision et d’intelligence artificielle (LIVIA) 
\\ École de technologie supérieure, Montreal, Canada}
\begin{document}
%
\maketitle
\begin{abstract}
In video-based emotion recognition (ER), it is important to effectively leverage the complementary relationship among audio (A) and visual (V) modalities, while retaining the intra-modal characteristics of individual modalities. In this paper, a recursive joint attention model is proposed along with long short-term memory (LSTM) modules for the fusion of vocal and facial expressions in regression-based ER. Specifically, we investigated the possibility of exploiting the complementary nature of A and V modalities using a joint cross-attention model in a recursive fashion with LSTMs to capture the intra-modal temporal dependencies within the same modalities as well as among the A-V feature representations. By integrating LSTMs with recursive joint cross-attention, our model can efficiently leverage both intra- and inter-modal relationships for the fusion of A and V modalities. The results of extensive experiments\footnote{The code is available on GitHub: \url{https://github.com/praveena2j/RecurrentJointAttentionwithLSTMs}.} 
performed on the challenging Affwild2 and Fatigue (private) datasets indicate that the proposed A-V fusion model can significantly outperform  state-of-art-methods.
\end{abstract}



\section{Introduction}
\label{sec:intro}
Automatic emotion recognition (ER)  is a challenging problem due to the complex and extremely diverse nature of expressions across individuals and cultures. In most of the real-world applications, emotions are exhibited over a wide range of emotional states besides the six basic categorical expressions - anger, disgust, fear, happy, sad, and surprise. For instance, emotional states can be expressed as intensities of fatigue, stress, and pain over discrete levels. Similarly, the wide range of continuous emotional states are often formulated as dimensional ER, where the diverse and complex human emotions are represented along the dimensions of valence and arousal. Valence denotes the range of continuous emotional states pertinent to pleasantness, spanning from being very sad (negative) to very happy (positive). Similarly, arousal spans the range of emotional states related to intensity, from being very passive (sleepiness) to extremely active (high excitement). In this paper, we have focused on developing a robust model for regression-based ER in valence-arousal space, as well as for fatigue.    

A and V modalities often carry complementary relationships among themselves, which is crucial to be exploited in order to build an efficient A-V fusion system for regression-based ER. In addition to the inter-modal relationships across A and V modalities, temporal dynamics in videos carry significant information pertinent to the evolution of facial and vocal expressions over time. Therefore, effectively leveraging both the inter-modal association across the A and V modalities and temporal dynamics (intra-modal) within A and V modalities plays a major role in building a robust A-V recognition system. In this paper, we have investigated the prospect of leveraging these inter- and intra-modal characteristics of A and V modalities in a unified framework. In most of the existing approaches for regression-based ER, LSTMs has been used in order to model the intra-modal temporal dynamics in videos \cite{cite6,9320301} due to their efficiency in capturing the long-term temporal dynamics \cite{9857074}. On the other hand, cross-attention models \cite{9667055} have been explored to model the inter-modal characteristics of A and V modalities for dimensional ER. 

In this work, we have proposed a unified framework for A-V fusion, which effectively leverages both the intra- and inter-modal information in videos using LSTMs and joint cross attention respectively. In order to further improve the A-V feature representations of the joint cross-attention model, we have also explored the recursive attention mechanism. Training the joint cross-attention model recursively allows refining the A and V feature representations, thereby improving the system performance. The main contributions of the paper are as follows. 
(1) A recursive joint cross-attentional model for A-V fusion is introduced to effectively exploit the complementary relationship across modalities while deploying a recursive mechanism to further refine the A-V feature representations. 
(2) LSTMs are further integrated to effectively capture the temporal dynamics within the individual modalities, as well as within the A-V feature representations. 
(3) An extensive set of experiments are conducted on the challenging Affwild2 and Fatigue (private) datasets, showing that the proposed A-V fusion model outperforms the related state-of-the-art models for regression-based ER.

\section{Related Work}
An early DL approach for A-V fusion-based dimensional ER was proposed by Tzirakis et al. \cite{cite7}, where the deep features (obtained with Resnet-50 for V and 1D CNN for A) are concatenated and fed to an LSTM. Recently, Vincent et al. \cite{9857074} investigated the effectiveness of attention models and compared them with recurrent networks. They have shown that LSTMs are quite efficient in capturing the temporal dependencies when compared to attention models for dimensional ER. Kuhnke et al. \cite{9320301} proposed a two-stream A-V network, where deep models are used to extract A and V features, and further concatenated for dimensional ER. Most of these approaches fail to effectively capture the intermodal semantics across A and V modalities. In \cite{cite8} and \cite{srini_2021_SLT}, authors focused on cross-modal attention using transformers to exploit the inter-modal relationships of A and V modalities for dimensional ER. Rajasekhar et al \cite{9667055} explored cross-attention models to leverage the inter-modal characteristics based on cross-correlation across the A and V features. They improved their approach by introducing joint feature representation into the cross-attention model to retain the intra-modal characteristics \cite{10005783, 9856650}. In most of these approaches, they cannot  effectively leverage intra-modal relationships. Chen et al. \cite{10.1145/2964284.2967286} modeled A and V features using LSTMs, and the unimodal predictions are combined using attention weights from conditional attention based on LSTMs. Priyasad et al. \cite{priyasad19_avsp} also explored LSTMs for V features, and used DNN-based attention on the concatenated features of A and V modalities for the final output predictions. Beard et al. \cite{beard-etal-2018-multi} proposed a recursive recurrent attention model, where LSTMs are augmented using an additional shared memory state in order to capture the multi-modal relationships in a recursive manner. In contrast with these approaches, we focus on modeling the A-V relationships by allowing the A and V features to interact and measure the semantic relevance across and within the modalities in a recursive fashion before feature concatenation. LSTMs are employed for temporal modeling of both uni-modal and multimodal features to further enhance the proposed framework. Therefore, our proposed model effectively leverages the intra- and complementary inter-modal relationships, resulting in a higher level of performance. 

\begin{figure*}[htb]
\centering
\centerline{\includegraphics[width=18cm]{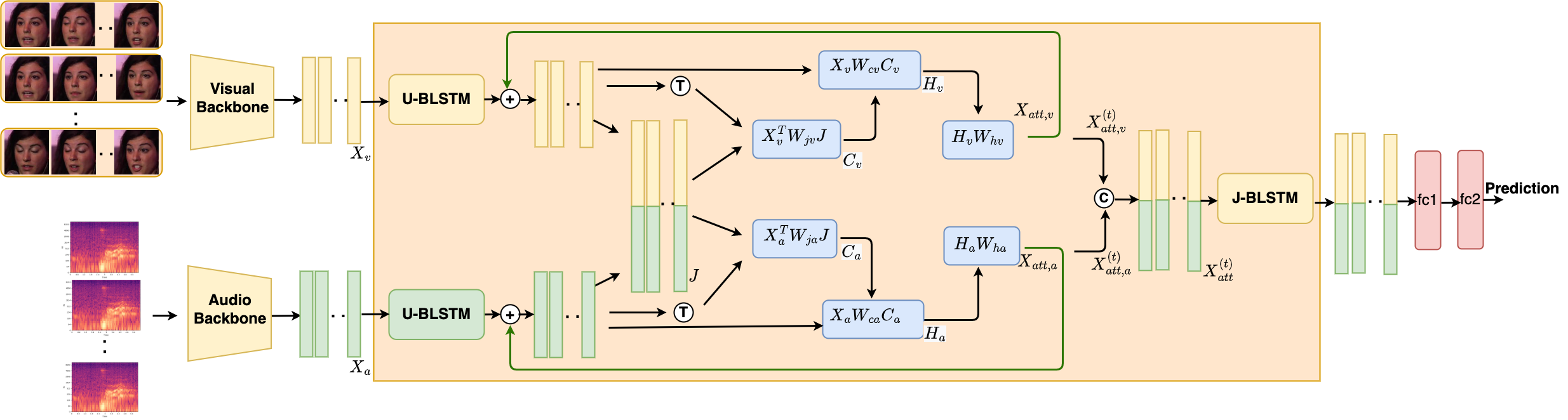}}
\caption{Block diagram of the proposed recursive joint attention model with BLSTMs.} \label{fig:res}
\end{figure*}

\section{Proposed Approach}

\noindent \textbf{A) Problem Formulation:}
Given an input video sub-sequence $S$, $L$ non-overlapping video clips are uniformly sampled and deep feature vectors ${\boldsymbol X}_{\mathbf a}$ and ${\boldsymbol X}_{\mathbf v}$ are extracted for the individual A and V modalities respectively from pre-trained networks. Let ${ \boldsymbol X}_{\mathbf a}=  \{ \boldsymbol x_{\mathbf a}^1, \boldsymbol x_{\mathbf a}^2, ..., \boldsymbol x_{\mathbf a}^L \} \in \mathbb{R}^{d_a\times L}$ and ${ \boldsymbol X}_{\mathbf v}=  \{ \boldsymbol x_{\mathbf v}^1, \boldsymbol x_{\mathbf v}^2, ..., \boldsymbol x_{\mathbf v}^L \} \in \mathbb{R}^{d_v \times L}$   
where ${d_a}$ and ${d_v}$ represent the dimensions of the A and V feature representations, respectively, and $\boldsymbol x_{\mathbf a}^{ l}$ and $\boldsymbol x_{\mathbf v}^{ l}$ denotes the A and V feature vectors of the video clips, respectively, for $l = 1, 2, ..., L$ clips. The objective of the problem is to estimate the regression model $F:\boldsymbol{X} \to \boldsymbol{Y}$ from the training data $\boldsymbol X$, where $\boldsymbol X$ denotes the set of A and V feature vectors of the input video clips and $\boldsymbol Y$ represents the regression labels of the corresponding video clips.

\noindent \textbf{B) Audio and Visual Networks:}
Spectrograms has been found to be promising with various 2D-CNNs (Resnet-18 \cite{7780459}) for ER \cite{10.1145/3428690.3429153,albanie}. Therefore, we  have explored spectrograms in the proposed framework. In order to effectively leverage the temporal dynamics within A modality, we have also explored LSTMs across the temporal segments of the A sequences. Finally, the A feature vectors of $L$ video clips are shown as      
$\boldsymbol{X_a} = ({\boldsymbol x_{\mathbf a}^1, \boldsymbol x_{\mathbf a}^2, ..., \boldsymbol x_{\mathbf a}^L})  \in\mathbb{R}^{{d_a}\times L}$.

Facial expressions exhibit significant information pertinent to both visual appearance and temporal dynamics in videos. LSTMs are found to be efficient in capturing the long-term temporal dynamics while 3DCNNs are effective in capturing the short-term temporal dynamics \cite{10.1145/2993148.2997632}. Therefore, we have used LSTMs with 3D CNNs (R3D \cite{8578773}) to obtain the V features for the fusion model. In most of the existing approaches, the output of the last convolution layer is 512 x 7 x 7, which is further passed through a pooling operation in order to reduce the spatial dimensions to 1 (7 $\to$ 1). This reduction in spatial dimension was found to leave out useful information as the stride is big. Therefore, inspired by the idea of \cite{9423042}, we use the A feature representation to smoothly reduce the spatial dimensions of raw V features for each video clip similar to that of \cite{9423042}. Finally, we obtain a matrix of V feature vectors of the video clips as
$\boldsymbol{X_v} = ({\boldsymbol x_{\mathbf v}^1, \boldsymbol x_{\mathbf v}^2, ..., \boldsymbol x_{\mathbf v}^L})  \in\mathbb{R}^{{d_v}\times L}$.  

\noindent \textbf{C) Recursive Joint Attention Model:}
Given the A and V features, $\boldsymbol{X_a}$ and $\boldsymbol{X_v}$, the joint feature representation is obtained by concatenating the A and V feature vectors  
$ {\boldsymbol J} = [{\boldsymbol X}_{\mathbf a} ; {\boldsymbol X}_{\mathbf v}] \in\mathbb{R}^{d\times L}$, 
where $d = {d_a} + {d_v}$ denotes the dimensionality of concatenated features. The concatenated A-V feature representations ($\boldsymbol J$) of a video sub-sequence ($\boldsymbol S$) are now used to attend to unimodal feature representations ${\boldsymbol X}_{\mathbf a}$ and ${\boldsymbol X}_{\mathbf v}$. The joint correlation matrix $\boldsymbol C_{\mathbf a}$ across the A features ${\boldsymbol X}_{\mathbf a}$, and the combined A-V features $\boldsymbol J$ are given by: 
\begin{equation}
   \boldsymbol C_{\mathbf a}= \tanh \left(\frac{{\boldsymbol X}_{\mathbf a}^T{\boldsymbol W}_{\mathbf j \mathbf a}{\boldsymbol J}}{\sqrt d}\right)
\end{equation}
where ${\boldsymbol W}_{\mathbf j \mathbf a} \in\mathbb{R}^{L\times L} $ represents learnable weight matrix across the A and combined A-V features, and $T$ denotes transpose operation. Similarly, the joint correlation matrix for V features are given by: 
\begin{equation}
   \boldsymbol C_{\mathbf v}= \tanh \left(\frac{{\boldsymbol X}_{\mathbf v}^T{\boldsymbol W}_{\mathbf j \mathbf v}{\boldsymbol J}}{\sqrt d}\right)
\end{equation}
The joint correlation matrices capture the semantic relevance across the A and V modalities as well as within the same modalities among consecutive video clips, which helps in effectively leveraging intra- and inter-modal relationships. After computing the joint correlation matrices, the attention weights of the A and V modalities are estimated. For the A modality, the joint correlation matrix $\boldsymbol C_{\mathbf a}$ and the corresponding A features ${\boldsymbol X}_{\mathbf a}$ are combined using the learnable weight matrices $\boldsymbol W_{\mathbf c \mathbf a}$ to compute the attention weights of A modality, which is given by 
$\boldsymbol H_{\mathbf a}=ReLU(\boldsymbol X_{\mathbf a} \boldsymbol W_{\mathbf c \mathbf a} {\boldsymbol C}_{\mathbf a})$
where ${\boldsymbol W}_{\mathbf c \mathbf a} \in\mathbb{R}^{{d_a}\times {d_a}} $ and ${\boldsymbol H}_{\mathbf a}$ represents the attention maps of the A modality. Similarly, the attention maps ($\boldsymbol H_{\mathbf v}$) of V modality are obtained as 
$\boldsymbol H_{\mathbf v}=ReLU(\boldsymbol X_{\mathbf v} \boldsymbol W_{\mathbf c \mathbf v} {\boldsymbol C}_{\mathbf v})$
where ${\boldsymbol W}_{\mathbf c \mathbf v} \in\mathbb{R}^{{d_v}\times {d_v}} $.
Then, the attention maps are used to compute the attended features of A and V modalities as: 
\begin{equation}
{\boldsymbol X}_{\mathbf a \mathbf t \mathbf t, \mathbf a} = \boldsymbol H_{\mathbf a} \boldsymbol W_{\mathbf h \mathbf a} + \boldsymbol X_{\mathbf a}
\end{equation}
\begin{equation}
{\boldsymbol X}_{\mathbf a \mathbf t \mathbf t, \mathbf v} = \boldsymbol H_{\mathbf v} \boldsymbol W_{\mathbf h \mathbf v} + \boldsymbol X_{\mathbf v}  
\end{equation}
where $\boldsymbol W_{\mathbf h \mathbf a} \in\mathbb{R}^{d\times {d_a}}$ and $\boldsymbol W_{\mathbf h \mathbf v} \in\mathbb{R}^{d\times {d_v}}$ denote the learnable weight matrices for A and V respectively. After obtaining the attended features they are fed again to the joint cross-attentional model to compute the new A and V feature representations as:
\begin{equation}
{\boldsymbol X}_{\mathbf a \mathbf t \mathbf t, \mathbf a}^{(t)} = \boldsymbol H_{\mathbf a}^{(t)} \boldsymbol W_{\mathbf h \mathbf a}^{(t)} + \boldsymbol X_{\mathbf a}^{(t-1)}
\end{equation}
\begin{equation}
{\boldsymbol X}_{\mathbf a \mathbf t \mathbf t, \mathbf v}^{(t)} = \boldsymbol H_{\mathbf v}^{(t)} \boldsymbol W_{\mathbf h \mathbf v}^{(t)} + \boldsymbol X_{\mathbf v}^{(t-1)}  
\end{equation}
where $\boldsymbol W_{\mathbf h \mathbf a}^{(t)} \in\mathbb{R}^{d\times {d_a}}$ and $\boldsymbol W_{\mathbf h \mathbf v}^{(t)} \in\mathbb{R}^{d\times {d_v}}$ denote the learnable weight matrices of $t^{th}$ iteration for A and V respectively.
Finally, the attended A and V features after $t$ iterations are further concatenated and fed to BLSTM to obtain the temporal dependencies within the refined A-V feature representations, which are fed to fully connected layers for final prediction. 

\section{Results and Discussion} 

\noindent \textbf{A) Datasets:}
Affwild2 is among the largest public dataset in affective computing, consisting of $564$ videos collected from YouTube, all captured in-the-wild \cite{Kollias}. 
The annotations are provided by four experts using a joystick and the final annotations are obtained as the average of the four raters. In total, there are $2,816,832$ frames with $455$ subjects, out of which $277$ are male and $178$ female. The annotations for valence and arousal are provided continuously in the range of $\lbrack-1,1\rbrack$. The dataset is 
partitioned in a subject-independent manner so that every subject’s data will be present in only one subset. The partitioning produces 341, 71, and 152 videos for the training, validation, and test sets respectively.

The Fatigue dataset (private) is obtained from $18$ participants in a Rehabilitation center, suffering from degenerative diseases inducing fatigue. A total of 27 video sessions are captured with a duration of 40 - 45 minutes and labeled at sequence level on a scale of 0 to 10 for every 10 to 15 minutes. We have considered $80\%$ of data as training data (50,845 samples) and $20\%$ as validation data (21,792 samples). 

\noindent \textbf{B) Ablation Study:}
Table \ref{Ablation Study} presents the results of the experiments conducted on the validation set for the ablation study. The performance of the approach is evaluated using Concordance Correlation Coefficient (CCC). In this section, we have analyzed the contribution of BLSTMs, where we have performed experiments with and without BLSTMs. Firstly, we have conducted experiments without using BLSTM for both the individual A and V representations as well as the A-V feature representations. Then, we included BLSTMs only for the individual A and V modalities before feeding to the joint attention fusion model i.e., only Unimodal-BLSTMs (U-BLSTMs). By including U-BLSTMs to capture the temporal dependencies within the individual modalities, we can observe improvements in performance. Therefore, BLSTMs are found to be promising in capturing the intra-modal temporal dynamics better than that of correlation-based intra-modeling in the joint attention model. After that, we have also included joint BLSTM (J-BLSTM) in order to capture the temporal dynamics across the joint A-V feature representations, which further improved the performance of the system. Note that in the above experiments, we have not performed recursive attention.  
In addition to the impact of U-BLSTM and J-BLSTMs, we have also conducted a few more experiments to investigate the impact of the recursive behavior of the joint attention model. First, we did recursion without LSTMs and found some improvement due to recursion. Then we included LSTMs and conducted several experiments by varying the number of recursions (iterations) in the fusion model. As we increase the number of recursive times, the model performance increases and starts to decrease after a certain recursion number. A similar trend of the model performance is also observed in the test set. Therefore, this can be attributed to the fact that recursion also works as a regularizer which improves the generalization ability of the model. In our experiments, we found that $t=2$ gives the best performance i.e, we have achieved the best performance of our model with two recursive iterations. 
\begin{table}[h]
\small
\centering
\caption{\textbf{Performance of our approach with components of BLSTM and recursive attention on Affwild2 dataset.}}
\label{Ablation Study}
\begin{tabular}{|l|c|c|c|c||c|c|c|c|c|c|} 
\hline
\textbf{Method}  & \textbf{Valence} &  \textbf{Arousal} \\
\hline \hline
 \multicolumn{3}{|c|}{\textbf{JA Fusion w/o Recursion}}  \\
\hline	
Fusion w/o U-BLSTM  & 0.670 & 0.590  \\
\hline
Fusion w/o J-BLSTM & 0.691 & 0.646 \\
\hline
Fusion w/ U-BLSTM and J-BLSTM & 0.715 &  0.688 \\
\hline
\multicolumn{3}{|c|}{\textbf{JA Fusion w/ Recursion}}  \\
\hline	
JA Fusion w/o BLSTMs, $t=2\
$ & 0.703 & 0.623  \\
\hline
JA Fusion with BLSTMs, $t=2$ & \textbf{0.721} & \textbf{0.694}  \\
\hline
JA Fusion with BLSTMs, $t=3$ & 0.706 &  0.652 \\
\hline
JA Fusion with BLSTMs, $t=4$ & 0.685 &  0.601 \\
\hline
\end{tabular}
\end{table}\\
\noindent \textbf{C) Comparison to State-of-art:}
Table \ref{Comparison with state-of-the-art for Affwild2 validation} shows our results against relevant state-of-the-art A-V fusion models on the Affwild2 dataset. The Affwild2 dataset has recently been widely used for Affective Behavior Analysis in-the-Wild (ABAW) challenges \cite{kollias2020analysing,Kollias_2021_ICCV}. Therefore, we compare our  approach with relevant state-of-art approaches in the ABAW challenges. Kuhnke et al \cite{9320301} used a simple feature concatenation using Resnet-18 for A and R3D for V modalities, showing better performance for arousal than valence. Zhang et al \cite{9607460} proposed a leader-follower attention model for fusion, and improved the performance (arousal) of the model proposed by Kuhnke et al \cite{9320301}. Rajasekhar et al. \cite{9667055} explored the cross-attention model by leveraging only the inter-modal relationships of A and V, and showed improvement for valence but not for arousal. Rajasekhar et al. \cite{10005783} further improved the performance of the model by introducing joint feature representation to the cross-attention model. The proposed model performs even better than that of vanilla JCA \cite{10005783} by introducing LSTMs as well as a recursive attention mechanism. 
\begin{table}
\footnotesize
\renewcommand{\arraystretch}{1.4}
\centering
\caption{ \textbf{CCC performance of the proposed and state-of-the-art methods for A-V fusion on the Affwild2 dataset.}}
    \label{Comparison with state-of-the-art for Affwild2 validation}
    \begin{tabular}{|l|c|c|c|c|c|c||c|c|c|} 
	\hline
	 \textbf{Method}   &  \textbf{Type of Fusion} &\textbf{Valence} & \textbf{Arousal}  \\ 
	 \hline  \hline
  \multicolumn{4}{|c|}{\textbf{Validation Set}}  \\
	\hline
    Kuhnke et al. \cite{9320301} & Feature Concatenation & 0.493 & 0.613 \\
	\hline
   Zhang et al. \cite{9607460}  & Leader Follower Attention & 0.469 & 0.649 \\
   \hline
	 Rajasekhar et al \cite{9667055} & Cross Attention & 0.541 & 0.517\\ 
	\hline 
    Rajasekhar et al. \cite{10005783} & Joint Cross Attention & 0.657 & 0.580\\ 
	\hline
       Ours & LSTM + Transformers & 0.628 & 0.654\\ 
	\hline
     Ours & Recursive JA + BLSTM & \textbf{0.721} & \textbf{0.694}\\
	\hline
   \multicolumn{4}{|c|}{\textbf{Test Set}}  \\
   \hline
      Meng et al. \cite{9857097} & LSTM + Transformers & 0.606 & 0.596 \\
	\hline
	Vincent et al. \cite{9857074} & LSTM + Transformers & 0.418 & 0.407 \\
	\hline
   Rajasekhar et al \cite{10005783} & Joint Cross Attention &0.451 & 0.389 \\
	\hline
    Ours & Recursive JA + BLSTM & \textbf{0.467} & \textbf{0.405}   \\
    \hline
\end{tabular}
\end{table} 
For test set results, the winner of the latest ABAW challenge \cite{9857097} has shown improvement using A-V fusion, however using three external datasets and multiple backbones. We have also compared the performance of our A and V backbones with the ensembling of LSTMs and transformers \cite{9857097} on the validation set. Vincent \cite{9857074} used LSTMs to capture intra-modal dependencies and explored transformers for cross-modal attention, however, they fail to effectively capture the inter-modal relationships across the consecutive video clips. Rajasekhar et al \cite{10005783} further improved the performance (valence) using joint cross-attention. The proposed model outperforms both \cite{9857074} and \cite{10005783}.  

Table \ref{Fatigue} shows the performance of the proposed approach on the Fatigue dataset. We have shown the performance of individual modalities along with feature concatenation and cross-attention \cite{9667055}. The proposed approach outperforms cross-attention \cite{9667055} and baseline feature concatenation.

\begin{table}
\small
\renewcommand{\arraystretch}{1.2}
    \centering
      \caption{ \textbf{CCC performance on the Fatigue dataset.}}
    \label{Fatigue}
\begin{tabular}{|l|c|c|c|c|c|c|c|c|c|c|} 
	\hline
	 \textbf{Method}  
	 & \textbf{Fatigue Level} \\
	 \hline \hline
	 Audio only (2D-CNN: Resnet18) & 0.312 \\
	\hline
    Visual only (3D-CNN: R3D) & 0.415\\
	\hline
	Feature Concatenation & 0.378 \\
	\hline
    Cross Attention \cite{9667055} & 0.421 \\
	\hline
    Recursive JA + BLSTM (Ours) & \textbf{0.447} \\
	\hline
\end{tabular}
\end{table}

\section{Conclusion}
This paper introduces a recursive joint attention model along with BLSTMs that allows effective spatiotemporal A-V fusion for regression-based ER. In particular, the joint attention model is trained in a recursive fashion, allowing to refine the A-V features.  We further investigated the impact of BLSTMs for capturing the intra-modal temporal dynamics of individual A and V modalities, as well as A-V features for regression-based ER. By effectively capturing the intra-modal relationships using BLSTMs, and inter-modal relationships using recursive joint attention, the proposed approach is able to outperform the related state-of-the-art approaches. 

 
\bibliographystyle{IEEEbib}
\bibliography{refs}

\end{document}